\documentclass[runningheads]{llncs}
\usepackage{graphicx}
\usepackage{subcaption}
\usepackage{amsmath}
\usepackage{todonotes}
\usepackage{array}
\usepackage{booktabs}

\begin{document}
\title{Counterfactual Multi-Agent Reinforcement Learning with Graph Convolution Communication
\thanks{This material is based upon work supported by the National Science Foundation under Grant No. CNS: 1650512. This
work was conducted in the Center for Visual and Decision Informatics, a National Science Foundation Industry/University Cooperative Research Center.}}
\author{Jianyu Su\inst{1} \and
Stephen Adams\inst{1} \and
Peter A. Beling\inst{1}}
% %
\authorrunning{Su et al.}
\institute{University of Virginia, 151 Engineer's Way, Charlottesville, VA, 22904, U.S.A \\
\email{\{js9wv, sca2c, pb3a\}@virginia.edu}\\}
\maketitle              % typeset the header of the contribution
\begin{abstract}
We consider a fully cooperative multi-agent system where agents cooperate to maximize a system's utility in a partial-observable environment.  We propose that multi-agent systems must have the ability to (1) communicate and understand the inter-plays between agents and (2) correctly distribute rewards based on an individual agent's contribution. In contrast, most work in this setting considers only one of the above abilities. In this study, we develop an architecture that allows for communication among agents and tailors the system's reward for each individual agent. Our architecture represents agent communication through graph convolution and applies an existing credit assignment structure, counterfactual multi-agent policy gradient (COMA), to assist agents to learn communication by back-propagation. The flexibility of the graph structure enables our method to be applicable to a variety of multi-agent systems, e.g. dynamic systems that consist of varying numbers of agents and static systems with a fixed number of agents. We evaluate our method on a range of tasks, demonstrating the advantage of marrying communication with credit assignment. In the experiments, our proposed method yields better performance than the state-of-art methods, including COMA. Moreover, we show that the communication strategies offers us insights and interpretability of the system's cooperative policies. 
\end{abstract}
\section{Introduction}

% \begin{itemize}
%     \item What is the general problem?
%         \begin{itemize}
%             \item Communication in multi-agent systems (MAS)
%         \end{itemize}
%     \item MARL is one way of modeling MAS
%         \begin{itemize}
%             \item What are the state-of-the-art methods?
%             \item What are the limitations of these methods?
%         \end{itemize}
%     \item Proposed method
%         \begin{itemize}
%             \item Describe method
%             \item How does it fill the gap in the literature?
%             \item How is it different from other methods?
%         \end{itemize}
%     \item What is our contribution?
% \end{itemize}

Communication, taking many forms in different scenarios, is closely associated with cooperation. For example, many species utilize vocal communication to serve different cooperative tasks e.g. mating, warning of predators etc. The ability to communicate can be vital for solving cooperative multi-agent reinforcement learning (MARL) problems, such as coordination of semi or full autonomous vehicles, and coordination of machines in a product line. By surveying the recent advances in the field of deep MARL, we identified two challenges in this domain.

% \todo{Before this paragraph, we should have a paragraph describing the current state of MARL} 
One challenge of MARL is the uncertainty of other agents' strategies through out the training process, making it hard for agents to understand the inter-plays and achieve cooperation. Communication methods emerged as solutions to overcome this challenge \cite{das2019tarmac,jiang2018learning,mordatch2018emergence,sukhbaatar2016learning}. However, information sharing among all agents could be problematic in that it poses difficulties for agents in learning valuable information from the large volume of shared information. In addition, in real-world applications, it can be expensive for all agents in the system to broadcast their information globally. Hence, recent communication frameworks have moved away from fixed communication protocols to flexible and targeted communication. The attentional communication model (ATOC) allows agents to select and communicate to ``collaborators'' \cite{jiang2018learning}. Targeted multi-agent communication (TarMAC) enables targeted communication between agents \cite{das2019tarmac}. 

Another key challenge is \textit{credit assignment}: without the full knowledge of the true state $s$ and other agent's actions, agents have issues deducing their contribution to the system's global reward $r$ \cite{foerster2018counterfactual}. It is possible to hand-engineer agent-level local reward functions in some scenarios \cite{gupta2017cooperative}, however, this technique has difficulties generalizing to other complex problems and requires domain expertise. Counterfactual multi-agent (COMA) policy gradient \cite{foerster2018counterfactual} has been proposed to solve such credit assignment problem in cooperative settings. COMA takes full advantages of centralised training that uses a centralized critic that conditions on the joint and all available state information and has achieved good performance on the \textit{StarCraft} environment. 

COMA addresses the credit assignment problem but overlooks communication between agents. The communication methods discussed above provide additional state information by information sharing but do not fully address the problem of credit assignment. In this paper, we propose a new multi-agent framework that can be seen as a factorization of a centralized policy that outputs actions for all the agents to optimize the expected global return. Combining communication with COMA, we allow agents to understand the inter-plays between agents and their contributions to the success of the system. Our framework consists of two components: a communication module that promotes an agent's understanding of inter-plays, and a credit assign framework that applies COMA to tailor the global reward to individual agents. We provide extensive empirical demonstration of the effectiveness and flexibility of our frame across a range of tasks, environments, and size of agents. We benchmark our framework on the traffic junction environment \cite{das2019tarmac,sukhbaatar2016learning} and show that agents are able to cooperate despite variations in team size. Further, through an investigation of communicated messages, we empirically demonstrate that meaningful communication strategies can be learned under the credit assignment training paradigm. Finally, we introduce a new manufacturing environment that consists of heterogeneous agents and requires intensive cooperation among agents to maximize the profits of the manufacturing system over a limited time. On empirical tests, our method outperforms state-of-art methods from the literature on this team task.

This paper is organized as follows. In Section \ref{sec:related}, we describe related work.  In Section \ref{sec:technical}, we introduce preliminary information about MARL and RL. We describe our proposed method in detail in Section \ref{sec:method}. In Section \ref{sec:exp}, we present the competitive performance on cooperative environment against baselines. We provide conclusions and possible areas of future work in Section \ref{sec:conclusions}. 

\section{Related Work}\label{sec:related}
\textbf{MARL} has benefited from recent developments in deep reinforcement learning, with the frameworks moving away from tabular methods \cite{bu2008comprehensive} to end-to-end schemes \cite{foerster2018counterfactual}. Our work is related to recent advances in deep multi-agent reinforcement learning. Judging by whether agents are allowed to communicate and whether the framework takes advantage of a centralized training paradigm, we categorize the recent deep multi-agent reinforcement learning works into communication methods and centralized training methods.

\textbf{Communication Methods}: This line of work utilizes an end-to-end communication architecture to enable agents to learn extra information from each other, where communication is learned by back-propagation. We can further categorize communication methods into explicit communication and implicit communication methods, based on whether the message is an input of agents. Explicit communication methods generally follow a sequential process where messages from agents at each timestep $t$ are generated, then messages are fed as inputs of other agents at $t+1$. \cite{foerster2016learning} defined discrete symbols as messages. \cite{das2019tarmac,jiang2018learning} expand communication vocabulary by utilizing continuous vector messages. 

Explicit communication methods might suffer from communication lag because messages only convey information about agents from the previous timestep. In contrast, implicit communication methods, integrating the message into the model's structure, allows for timely information sharing among agents. \cite{mordatch2018emergence} utilizes a communication module to generate discrete communication features for the final policy layer that outputs agents' actions. CommNet \cite{sukhbaatar2016learning} and BiCNet \cite{peng2017multiagent} communicate the encoding of local observation among all agents. Similar to our communication module, DGN utilizes graph convolution to allow for targeted communication between agents \cite{jiang2018graph}. 
Recent advances in communication methods, including our architecture, are often coupled with the utilization of \textit{self-attention}, which is referred to as a relation kernel. Among those works, TarMAC falls in the explicit communication category \cite{das2019tarmac}, and ATOC, which is categorized as a implicit communication method, does not use graph convolution. The closest to our communication module is DGN \cite{jiang2018graph}, which also utilizes graph convolutions and \textit{self-attention}. However, DGN considers a different problem setting where local rewards are presented to agents, while our method  provides only global rewards. Our method is further distinguished by the use of a recurrent structure in the policy network to take into consideration temporal factors.

\textbf{Centralized Training Methods}: Although communication methods also follow a centralized training paradigm, most of them are excluded from this category because they do not allow for true state information that helps individual agents to condition environment rewards on true states, thus fails to take full advantages of the training paradigm. Centralized training methods often use critics to promote an agent's understanding of their joint contribution to the overall system. \cite{gupta2017cooperative} utilizes a critic that conditions on an agent's local information and hand-crafted local reward. \cite{das2019tarmac,foerster2018counterfactual,lowe2017multi} use a centralised critic to evaluate the action-value of the joint actions. Among these methods, only COMA addresses the credit assignment problem. In the proposed method, we utilize the COMA framework to promote effective communication between agents by back-propagating the gradient evaluated by the central critic.

\section{Technical Background}\label{sec:technical}
\textbf{Decentralized Partially Observable Markov Decision Processes (Dec-POMDPs)}: 
Consider a fully cooperative multi-agent task with $N$ agents. The environment has a true state $s \in S$, a transition probability function $T$, and a global reward function $R$. Each agent takes actions $a_n \in A$ according to its own policy $\pi_{\theta_n}(a_n|o_n)$ with $n \in \{1,\cdots,N\}$, and receives partial state information in the form of observation $o_n \in O_n$. At each timestep, each agent simultaneously chooses an action $a_n$, forming a joint action $\mathbf{a}$ that induces a transition in the environment according to $T(s'|s, \mathbf{a})$.
The joint action $\mathbf{a}_t$, determined by the joint policy $\mathbf{\pi}$, induces an overall environment reward $r=R(s, \mathbf{a})$. We denote the joint quantities in bold, and use $\mathbf{a}^{-a_n}$ to denote the joint actions other than the agent $n$'s action $a_n$. Similar to a\ single-agent RL, MARL aims to maximize the discounted return $R_t=\sum_{l=1}^{\infty}\gamma ^l r_{t+l}$. The joint value function $V^{\mathbf{\pi}}(s_t)=E[R_t|s_t=s]$ is the expected return for following the joint policy $\mathbf{\pi}$ from state $s$. The value-action function $Q^{\pi}(s,\mathbf{a})  =E[R_t|s_t=s,\mathbf{a}]$ defines the expected return for selecting joint action $\mathbf{a}$ in state $s$ and following the joint policy $\mathbf{\pi}$.

\textbf{Deep Q-Network (DQN)}: Combining reinforcement learning with deep neural networks, DQN demonstrated human-level control policies in many Atari games \cite{mnih2015human}. DQN aims at updating its parameters by minimizing a sequence of loss functions at each iteration $i$: $L(\theta)_i=E_{(s,a,r,s')}[y_{i}'-Q^{\pi}(s,a;\theta_i)]$, where $y_i'=r+ \max_{a'} Q^{\pi}(a',s';\theta_{i-1})$. DQN is an \textit{off-policy} algorithm that takes raw pixels as input and outputs actions. It utilizes a replay buffer to store its past experience in tuples $(s,a,r,s')$, which allows for efficient sampling and resolving the strong correlation between consecutive samples.

\textbf{Independent Q-Learning (IQL)}: Similar to DQN which learns an action-value approximator $Q^{\pi}(s,a;\theta)$, IQL learns to approximate $Q^{\pi_n}(o_n, a_n;\theta_n)$ with respect to each agent in the multi-agent setting \cite{tan1993multi}. \cite{foerster2018counterfactual} employed a parameter sharing technique to modify this model to approximate $Q^{\pi_n}(o_n, a_n, id_n;\theta)$, where $id_n$ is a feature that used to differentiate agents. 
% IQL parameters are optimized by minimizing: $$E_{(s,\mathbf{a},r,s')}[y_{i}'-Q^{\mathbf{\pi}}(s,\mathbf{a};\theta_i)]=E_{n\sim N}\big[E_{(o_n,a_n,r,o_n')}[y_{i}'-Q^{\pi_n}(o_n, a_n, id_n;\theta_i)]\big],$$ where $y_i'$ is the target global return. 
% We argue that DGN, which is optimized by minimizing $E_{n\sim N}\big[E_{(o_n,a_n,r,o_n')}[y_{i}'(id_n)-Q^{\pi_n}(o_n, a_n, id_n;\theta_i)]\big]$, is a special case of IQL with individual target return $y_{i}'(id_n)$ available. \todo{read DGN and counterfactual paper}

\textbf{Policy Gradient Algorithms}: These algorithms directly adjust the parameters $\theta$ of the policy in order to maximize the objective $J(\theta)=E_{s\sim p^{\pi},a\sim \pi}[R(s,a)]$ by taking steps in the direction of $\nabla J(\theta)$. The gradient with respect to the policy parameters is $\nabla_{\theta} J(\theta)=E_{s\sim p^{\pi},a\sim \pi}[\nabla_{\theta} \log\pi_{\theta}(a|s)Q_{\pi}(s,a)]$, where $p^{\pi}$ is the state transition by following policy $\pi$, and $Q_{\pi}(s,a)$ is an action-value. Policy gradient algorithms differ in how to evaluate $Q_{\pi}(s,a)$, e.g. the REINFORCE algorithm \cite{williams1992simple} simply uses a sample return $R=\sum_{i=t}^T \gamma_{i-t}r_i$.
% Alternatively, one can introduce an action-value function, a \textit{critic}, to approximate the true action-value $Q_{\pi}(s,a)$ independently from the policy function $\pi$. This leads to a number of \textit{actor-critic} algorithms e.g. DDPG \cite{}.

\textbf{Multi-Agent Policy Gradient Algorithms}: Multi-agent policy gradient methods are extensions of policy gradient algorithms with a policy $\pi_{\theta_n}(a_n|o_n), n \in 1,\cdots,N$ for each agent. Similar to RL, many multi-agent implementations utilize critics to reduce the gradient variances. Among those multi-agent critic algorithms, COMA stands out as it deals with credit assignment issues. In the proposed method, we use the COMA structure to assist agents in learning to communicate by back-propagating tailored gradient. Details of COMA will be presented in the next section.

\section{Methods}\label{sec:method}
\begin{figure}[ht]
\centering
\includegraphics[width=.9\textwidth]{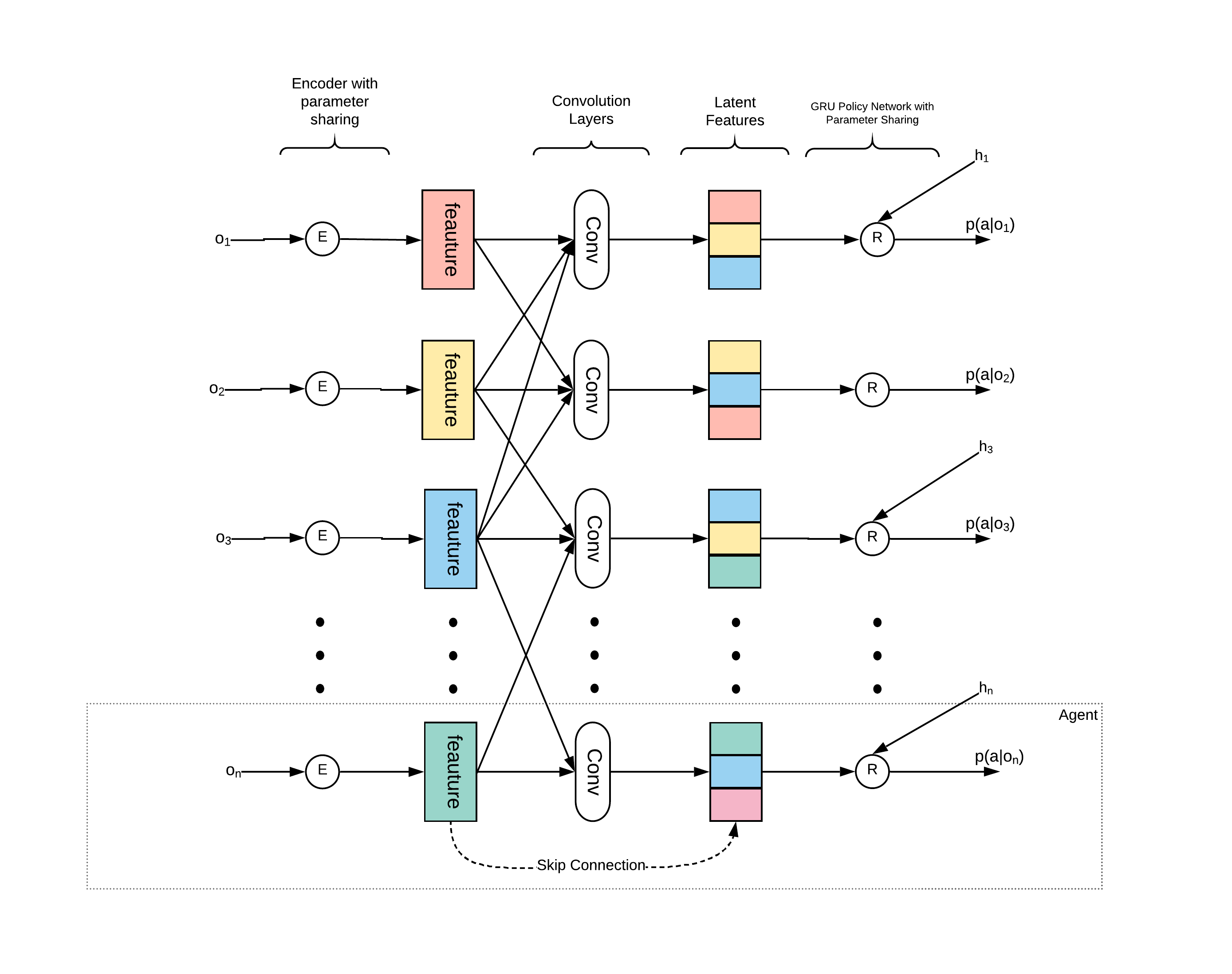}
\caption{Graph convolution integrates information from neighbouring nodes for Central nodes. This process is deemed as information sharing between neighbouring and central nodes.} \label{fig: gc}
\end{figure}

\begin{figure}[ht]
\centering
\includegraphics[width=.6\textwidth]{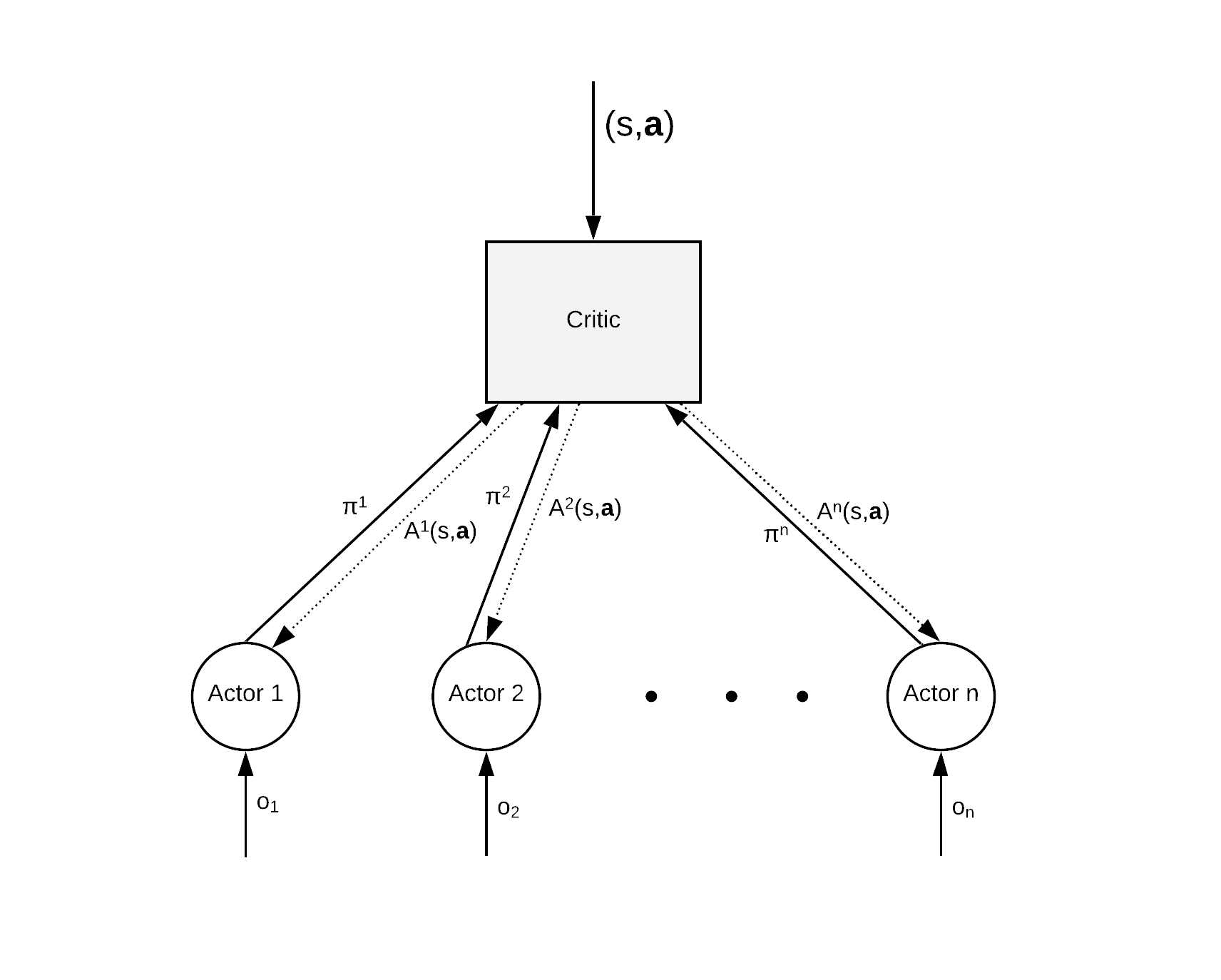}
\caption{Centralized training paradigm proposed in \cite{foerster2018counterfactual}.} \label{fig: centralized}
\end{figure}

 As shown in Figure \ref{fig: centralized}, our architecture can be described as a multi-agent actor-critic framework. In the \textbf{Dec-POMDP} environment we described in the previous section, the centralized critic has access to true states $s \in S$ (concatenations of local observations if true states are not available) while actors only receive local partial observation $o$. The communication module, illustrated in Figure \ref{fig: gc}, is embedded along the actor network and communication strategies are obtained by back-propagation without supervision. During training, the critic helps actors to learn how to communicate and condition their individual actions on the true states. During testing, actors execute without the presence of the critic. 
 
\subsection{Communication Module}
We now describe our multi-agent communication architecture in details. Our communication kernel consists of graph convolutions and relation kernels.

\textbf{Graph Convolution}:  Graph convolution extends the application of convolution neural networks from the euclidean domain to the non-euclidean domain such as graphs \cite{wu2019comprehensive}. Specifically, the spectral filter slides from one central node to another to learn latent features from clusters of nodes that are pivoted around the central node. Whenever a node is encoded into latent vectors, the aggregated information from its neighbouring nodes is taken into account by the spectral filter. Hence, we represent the communication among nodes with graph convolutions. In this work, we represent the environment as a graph, where each agent is represented by a node in the graph and edges are defined by metrics that measure the relationship (closeness) between agents. Agents can communicate with each other if and only if their corresponding nodes are connected in the graph. Moreover, the graph configuration may vary over time based on the metrics defined by domain experts, which makes our communication module applicable to dynamic environments.
% The intuition behind this is that agents that are closely related might affect and need to learn from each other. In addition, in complex environments, it may be costly and trivial to take all other agents into consideration, because receiving a large amount of information might pose problems to agents such as high computational complexity and difficulties to differentiate valuable information from the large volume of shared messages.

The communication module consists of three components: observation encoder, convolution layer, and an RNN that outputs multinomial policies. Let agent $n$ be an arbitrary agent in the environment, $m\in N$ indexes agents that are connected to agent $n$ in the graph. At timestep $t$, local observations, $o_n$ and $o_m$, are encoded into feature vectors $h_n$ and $h_m$ by the observation encoder. Then, the convolution layer aggregates $h_{n \cup m}$ from the neighbourhood of agent $n$. This \textit{one-hop} convolution process allows for agent $n$ to communicate with its first-order neighbours. By stacking $l$ convolution layers, agent $n$ can aggregate information from nodes that are up to $l$-hops away. The convolution process can be written as:
\begin{equation} \label{eq: convolution}
    h^{l+1}_{n}=\sigma \left( \sum_{j\in (m\cup n)} \alpha_{nj}h^l_{j}W^l \right).
\end{equation}
% \todo{Previously, we were not numbering equations.  Check ECML guidelines to see if all equations should be numbered.}
Here, $\sigma$ denotes an activation function, $h^{l+1}_{n}$ is agent $n$'s latent features generated by the $l$th convolution layer, $W^l$ represents trainable weights in the convolution layer, and $\alpha_{nm}$ is a relation weights that defines the amount of information agent $n$ takes from agent $m$.  A range of relation kernels has been proposed \cite{hamilton2017inductive,kipf2016semi,velivckovic2017graph}. The RNN takes latent features from convolution layers as the input and outputs agent's policy $\pi_n$.

\textbf{Relation Kernel}: Inspired by \cite{jiang2018graph}, we use multi-head dot-product attention (MHDPA) as the relation kernel to perform inductive reasoning between nodes. MHDPA projects the input matrix to a set of matrices that represent query, key, and value. Relation weights are obtained by dot-product attention. Formally, relation weights in the $l$th convolution layer, $\alpha^l_{nj}$, can be written as:
\begin{equation} \label{eq: attention}
    \alpha^l_{nm} = \frac{\exp (h_n^lW_Q^l (h_m^lW_K^l)^{T} / \sqrt{d_k})}{\sum_{k\in (m \cup n)} \exp (h_n^lW_Q^l (h_k^lW_K^l)^{T} / \sqrt{d_k})},
\end{equation}
 where $W_Q^l \in R^{d_k^{\prime}\times d_k}$ and $W_K^l \in R^{d_k^{\prime}\times d_k}$ are trainable weights that project inputs of size $N \times d_k^{\prime}$ into query matrices and key matrices, respectively, $h^l$ denotes input vectors with $n$ indexes central nodes, $m$ indexes neighbouring nodes, $d_k$ is the dimension of the $l$-layer latent feature dimension, and $d_k^{\prime}$ denotes input dimension.
By combining Equation \ref{eq: convolution} with \ref{eq: attention}, the latent features generated by the $l$th convolution layer can be rewritten as:
\begin{equation}
    h^{l+1}_{n}=\sigma \left(\sum_{j\in (m\cup n)} \alpha^l_{nj}h^l_{j}W^l_V\right),
\end{equation}
where $W_V^l \in R^{d_k^{\prime}\times d_k}$ are trainable weights that projects inputs into a value matrices.

Depending on the application, the underlying graph $G_t$ that depicts the relationship between agents might change over time, making it difficult to achieve flexible and efficient graph convolution implementations. In this work, we utilize a mask $M_t$ to filter out nodes that are not related, where $M^t=A^t+I$, with $A^t \in R^{N \times N}$ as the adjacency matrix at time step $t$, and $I \in R^{N \times N}$ as an identity matrix. Then $\alpha^l \in R^{N \times N}$ in the $l$th graph convolution layer can be written in the following matrix form:
\begin{equation} \label{eq: mask_attention}
    \alpha^l_{nj} = \frac{\exp (h_n^lW_Q^l (h_j^lW_K^l)^{T} \otimes \hat{M^t}_{nj}/ \sqrt{d_k})}{\sum_{k \in N} \exp (h_n^lW_Q^l (h_k^lW_K^l)^{T} \otimes \hat{M^t}_{nk}/ \sqrt{d_k})},
\end{equation}
where $\otimes$ denotes matrix element-wise product, $\hat{M^t} = (1 - M^t) \otimes (\delta)$ with $\delta$ being a small positive scalar, $n$ indexes the central nodes and $j$ indexes a random node in the graph.
\begin{figure}[ht]
\centering
\includegraphics[width=.5\textwidth]{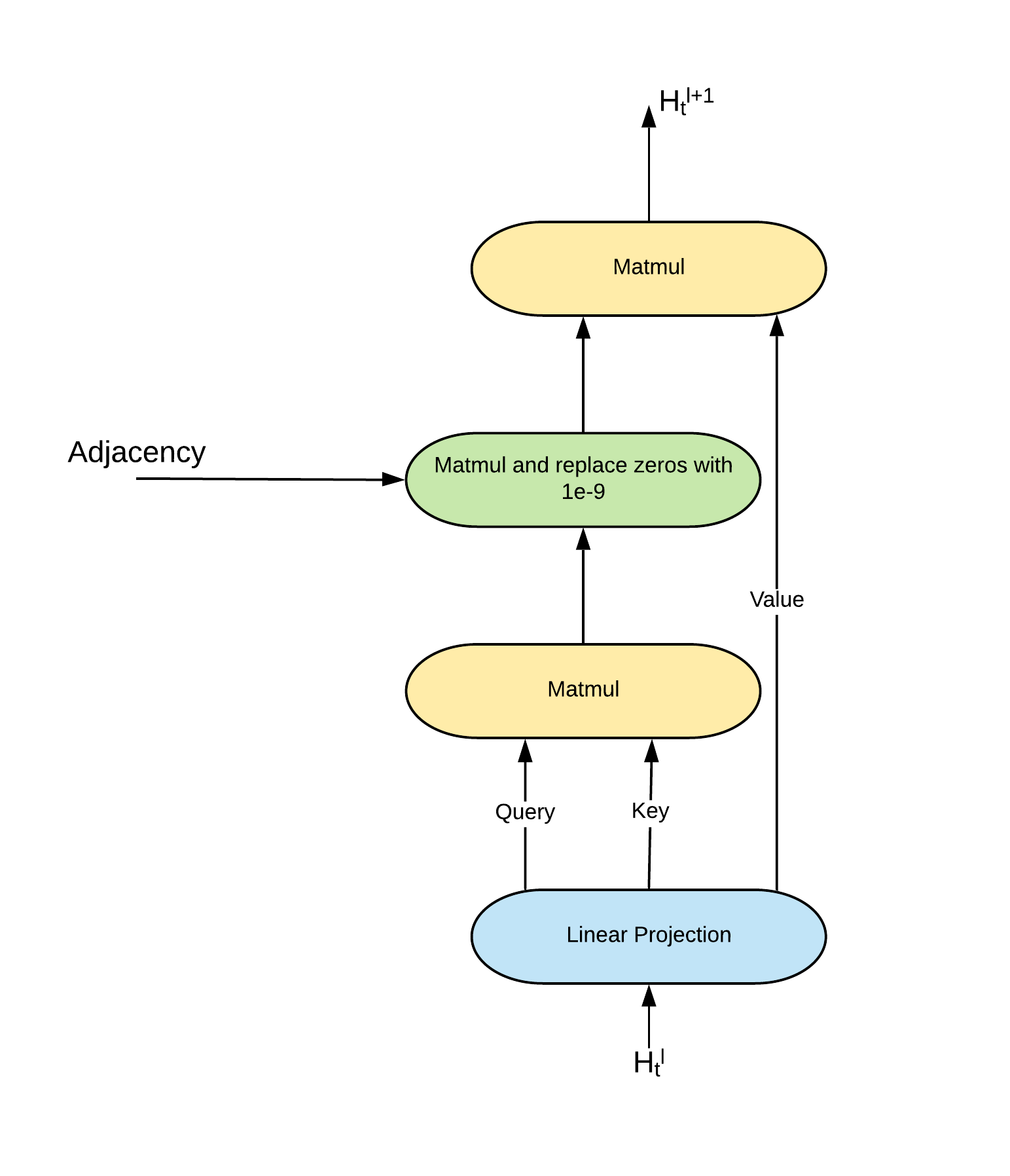}
\caption{Relational Kernel with flexible graph configurations} \label{fig: relation}
\end{figure}
\subsection{Training Paradigm}
In the MARL environment, the reward $r=R(s, \mathbf{a})$ conditions on true state $s$ and joint actions $\mathbf{a}$ rather than individual observation $o_n$ and actions $a_n$. The global reward $r$ does not reveal each agent's contribution, and thus the gradient calculated from the global rewards, such as $\nabla_{\theta_n}\log (a_n|o_n)r$, does not necessarily encourage individual agent $n$ to choose actions for the greater good of the whole system. As pointed out by \cite{wolpert2002optimal}, one has to conduct counterfactual experiments to find out the contribution from individual agents. That is, given a state $s_t$, we change agent $n$'s action $a_n$ while holding other agents' actions constant. Then we evaluate how the reward changes for agent $n$. \textit{Difference rewards} $D^{n}=R(s,\mathbf{a})-R(s,(\mathbf{a}^{-a_n}, a_n^c))$, which are developed based on this counterfactual reasoning, reveals the contribution from agent $n$ on the overall success of the system via comparing the global reward changes incurred by changing agent $n$'s actions to a default action $a_n^c$. Any action by agent $a$ that improves $D^{a}$ improves the global reward $r(s, \mathbf{a})$. 

Inspired by \textit{difference rewards}, COMA utilizes a centralized critic to evaluate the counterfactual rewards $Q(s,(\mathbf{a}^{-a_n}, a^{'}_n))$ for each agent instead of rolling out $|A|^{n}$ parallel simulations at each timestep. This renders the counterfactual advantage:
\begin{equation}
    A^{n}(s,\mathbf{a})=Q(s,\mathbf{a})-\sum_{a^{'}_n} \pi_{n}(a^{'}_n|o_n)Q(s,(\mathbf{a}^{-a_n},a^{'}_n)),
\end{equation}
where $A^{s, \mathbf{a}}$ is the tailored advantage for agent $n$, and $Q(s,\mathbf{a})$ is the action-value of the joint action evaluated by the centralized critic. The second part on the right-hand side is the separate counterfactual baseline marginalized on agent $n$'s available actions with $\pi_n(a^{'}_n|o_n)$ action probabilities outputted by the decentralized actor. 

We follow COMA's training paradigm to further promote agents' understanding of the inter-plays, with each agent obtaining tailored gradients (assuming each agent's policy is parameterized by $\theta_n$), $\nabla_{\theta_n} \log(a_n|o_n)A^n(s,\mathbf{a})$, from the centralized critic. Coupled with the communication module above, the actor network's gradient is given by:
\begin{equation}
   g_{\theta} =E_{\mathbf{\pi}} \bigg[\sum_{n} \nabla_{\theta} \log(a_n|o_n)A^n(s,\mathbf{a})\bigg],
\end{equation}
where $\theta$ parameterizes the actor network. The critic minimizes the temporal difference (TD) error:
\begin{equation}
    e=(1-\lambda)\sum_{n=1}^{\infty} \lambda^{n-1} (\sum_{l=1}^{n} \gamma^{l-1}r_{t+l}+\gamma^nQ(s_{t+n},\mathbf{a}_{t+n}))-Q(s_t,\mathbf{a}_t).
\end{equation}

Our implementation employs the A2C framework \cite{mnih2016asynchronous} to enable efficient on-policy sampling. That is, our model executes in multiple parallel environments, within which, the temporary buffer collects experiences from the current run $(s,\mathbf{a},\{o_1,o_2,\cdots,o_n\}, r)$, where $s$ denotes true states of the environment, $\mathbf{a}$ denotes the joint actions, $\{o^1,o^i,\cdots,o^n\}$ are local observations from $n$ agents respectively, and $r$ is the environment reward. 

% I need to put critic structure, critic loss here

% I need to put actor structure, actor loss here

\section{Experiments}\label{sec:exp}
To evaluate the effectiveness of the idea of combining communication with individual agent reward shaping, we compare our models with two baselines: COMA and IQL with proposed communication module. In this section, these three models are evaluated on 2 environments. The following subsections will present environment descriptions, model configuration, evaluations, and ablations. 

All models used in the following section share the same configuration: RMSProp with a learning rate of $5 \times 10^{-4}$ and $\alpha=0.99$, batch size $8$, discount factor $\gamma=0.99$ and trace decay discount TD$(\lambda)=0.8$. All methods containing communication modules utilize $2$-layer convolutions with $8$-head dot product attention.
 
\begin{table}[ht]
\centering
\caption{Summaries of models tested in the experiments}\label{tabel: baselines}
\begin{tabular}[t]{l>{\centering}p{0.22\linewidth}>{\centering\arraybackslash}p{0.22\linewidth}>{\centering\arraybackslash}p{0.22\linewidth}}
\toprule
&Graph Convolution  & Counterfactual Baselines & Description\\
\midrule
IQL with Communication & Yes & No & No reward shaping\\
COMA  & No & Yes & No communication\\
CCOMA  & Yes & Yes & Our architecture\\
\bottomrule
\end{tabular}
\end{table}

\subsection{Traffic Junction}

The traffic junction environment consists of vehicles in the environment moving along pre-assigned routes and was originally introduced in \cite{sukhbaatar2016learning}. At each timestep, new cars enter the grid with probability $p_{arrive}$ at designated starting positions. However, the total number of cars at any time is constraint by $N_{max}$. Each car's task is to move along the pre-assigned route without colliding with other cars. At every timestep, a car can either \textit{gas} which advances it by one cell on its route or \textit{break} to stay at its current grid. A car is removed from the environment once it reaches its designated goal.

Two cars \textit{collide} if they enter the same grid. This does not affect the simulation other than incurring the reward $r_{coll}$. The simulation also utilize $\tau r_{time}$ to discourage traffic jam at each timestep, where $r_{time} < 0$ is a reward and $\tau$ is the number of timesteps since the car entered the grid. Therefore, the total reward at time $t$ can be written as:
\begin{equation}
    r(t) = C^{t}r_{coll} + \sum_{i=1}^{N^t}\tau_{i}r_{time}, \nonumber
\end{equation}
where $C^{t}$ is the number of collisions at timestep $t$, and $N^t$ is the number of cars present. The game terminates after $40$ steps and is deemed as a failure if one or more collisions have occurred.

\begin{figure}[ht]

\begin{subfigure}{0.45\textwidth}
\centering
\includegraphics[width=0.6\linewidth]{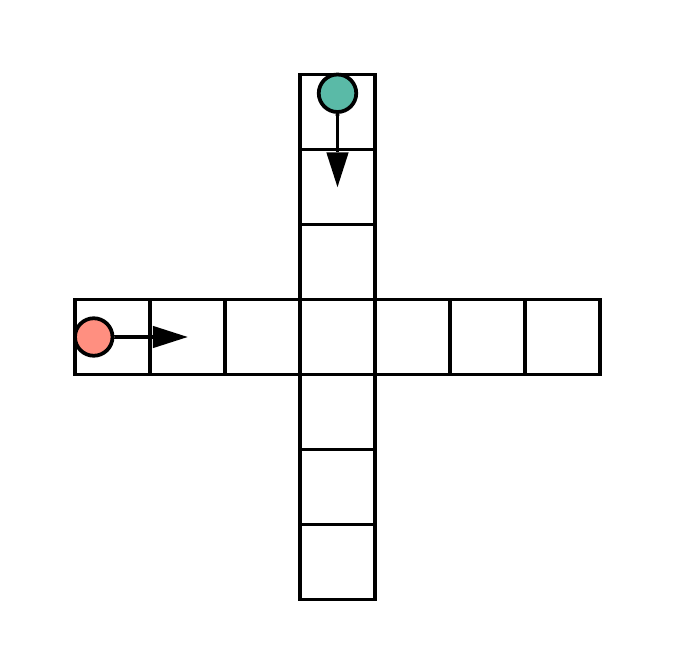} 
\caption{Easy mode - $N_{max}=5$ and $p_{arrive}=0.30$.}
\label{fig:subim1}
\end{subfigure}
\hfill
\begin{subfigure}{0.45\textwidth}
\includegraphics[width=0.85\linewidth]{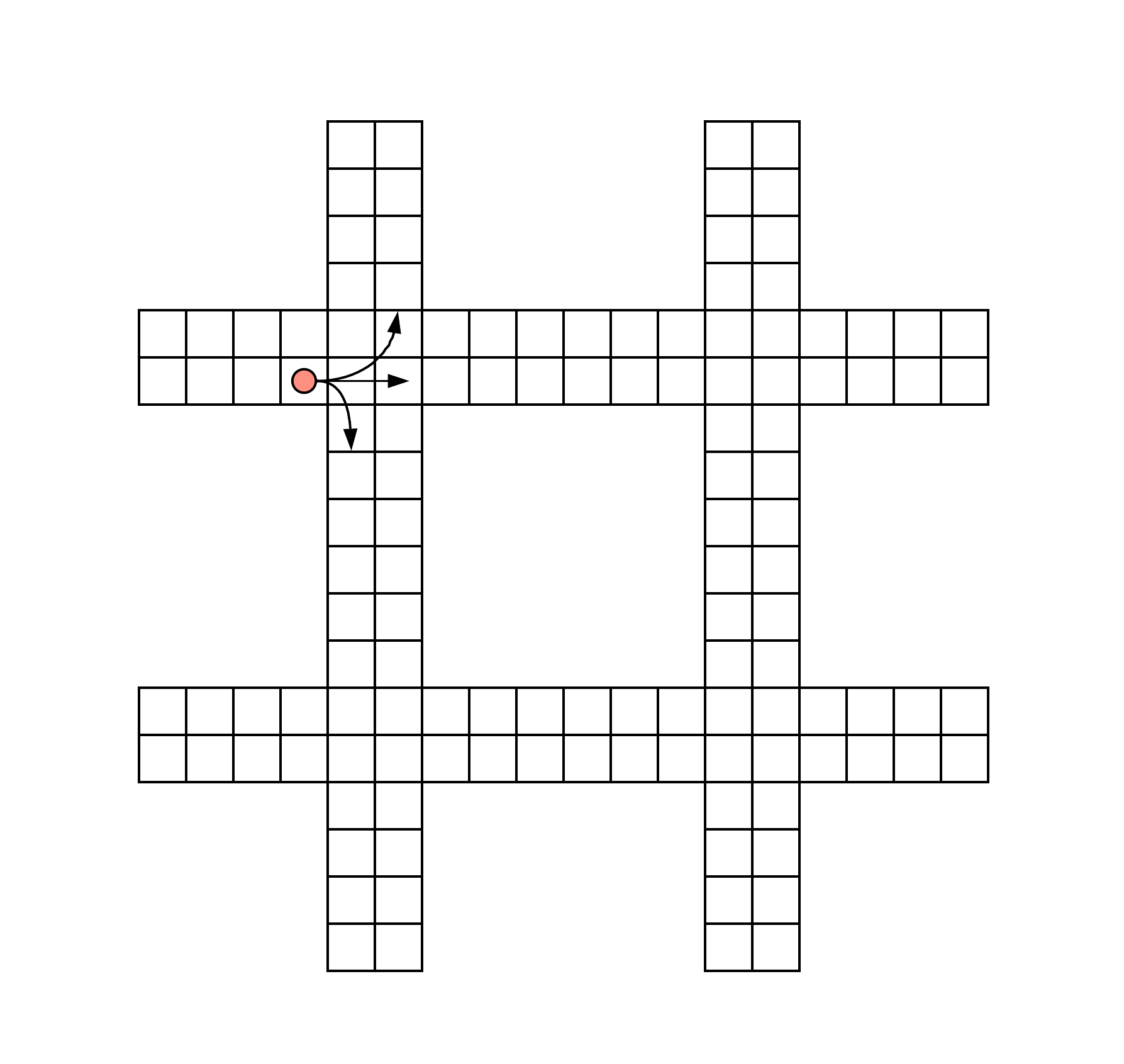}
\caption{Hard mode - $N_{max}=20$ and $p_{arrive}=0.05$.}
\label{fig:subim2}
\end{subfigure}

\caption{Traffic junction environment with two modes. The game continues for 40 steps. It is deemed a failure if more than one collisions occurs.}
\label{fig:tj_over}
\end{figure}

There are two modes of the simulation: easy and hard. In the easy mode, the environment is a $7 \times 7$ grid that contains one junction intersected by two one-way road as shown in Figure \ref{fig:subim1}. $N_{max}=5$ and $p_{arrive}=0.30$. The hard mode traffic junction environments consist of 4 junctions as shown in Figure \ref{fig:subim2}. At every timestep, cars can enter the grid from 1 of the 8 arrival points. And there are 7 different routes for each arrival points. The hard mode environment has $18 \times 18$ grid with $N_{max}=20$ and $p_{arrive}=0.05$..

All agents are represented by a tuple $\{id, position, path \}$, and they can only observe other cars in its vision range (a surrounding $3 \times 3$ neighbour). Thus, the input size for each agent's network is $9 \times 3$. For the centralized critic, we use a vector that tracks ids of active vehicles in the simulation to denote the true state of the environment. We trained all methods for $2$ million training steps, and optimal policies are saved. 

Table \ref{table: tj} shows the performance of each method on both easy mode and hard mode. The performance of our models in both modes outperforms CommNet and TarMAC reported in \cite{das2019tarmac}. 
% The agent in our method uses self-attention to reason \todo{what does reason the information mean?} the information from each source while agents of CommNet simply takes an average of received information. Although TarMAC also applies self-attention to generate messages, the messages encapsulate state information from all previous time steps. 
The comparison also demonstrates that our proposed communication module is effective with or without a centralized critic. Note that although COMA agents do not communicate with each other, the centralized critic is able to help agents that cannot communicate learn good policies.

\begin{table}[t] 
\centering
\caption{Success rates on the traffic junction task.}
\begin{tabular}{l l l l l l c}
\hline
 &  & &Easy & & Hard & Harder\\
\hline
IQL with comm & & &100.0\% & &98.6\% &94.3\%\\
COMA & & &100.0\% & &99.1\%&99.1\%\\
CCOMA & & &100.0\% & &99.6\%& 99.3\%\\
\hline
CommNet reported in \cite{das2019tarmac} & & & 99.7\% & &78.9\% & -\\
TarMAC reported in \cite{das2019tarmac} & & & 100.0\% & & 97.1\% & -\\
\hline
\label{table: tj}
\end{tabular}
\end{table}

\begin{figure}[t]
\begin{subfigure}{0.45\textwidth}
\centering
\includegraphics[width=0.8\linewidth]{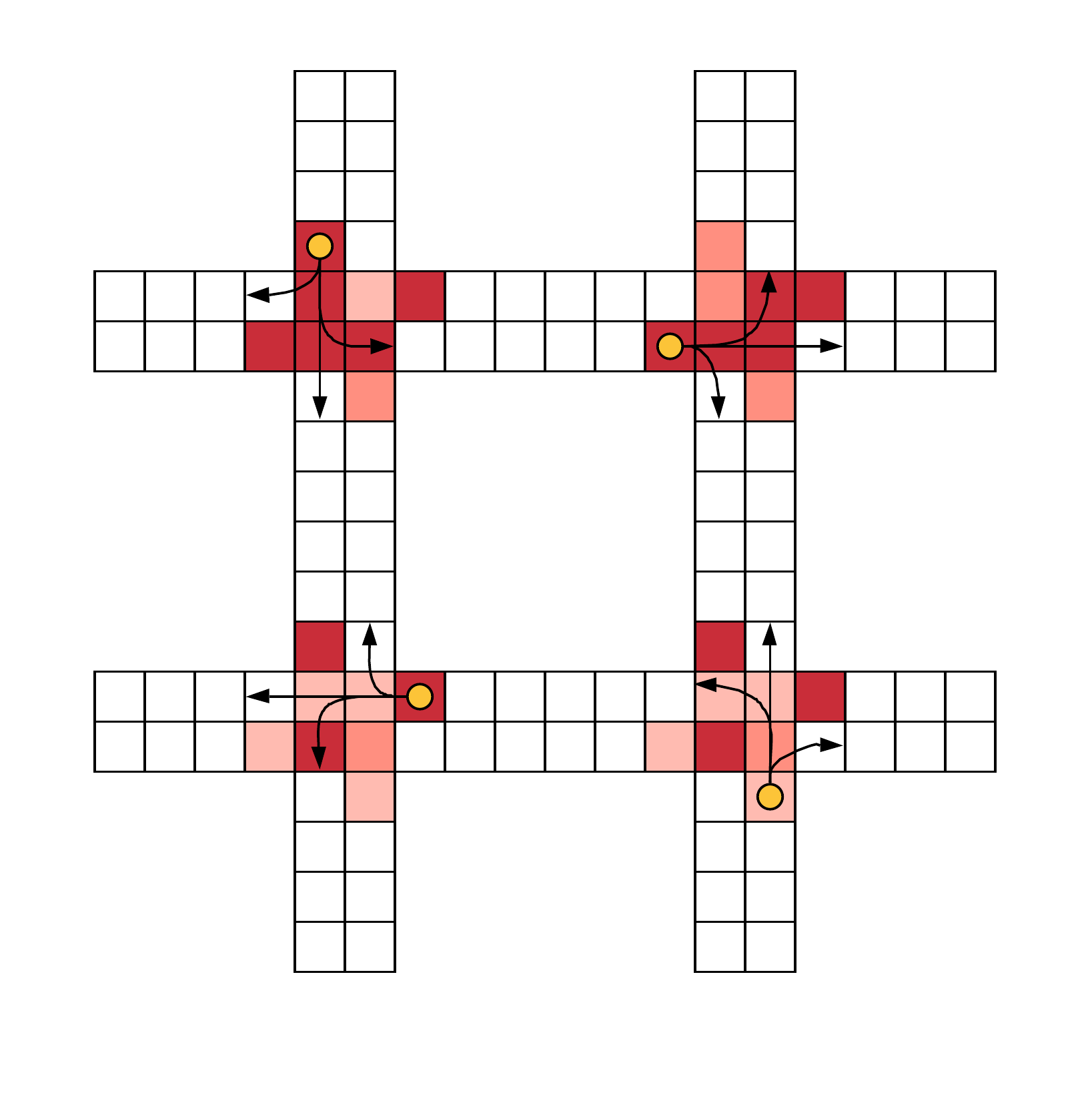} 
\caption{Most of brakes happen when vehicles are entering junctions or in the cell where vehicles have probabilities to turn. Once the vehicle across the junction the brake probabilities drop as well. Examples of vehicles' trajectories are shown.}
\label{fig:brake}
\end{subfigure}
\hfill
\begin{subfigure}{0.45\textwidth}
\includegraphics[width=1\linewidth]{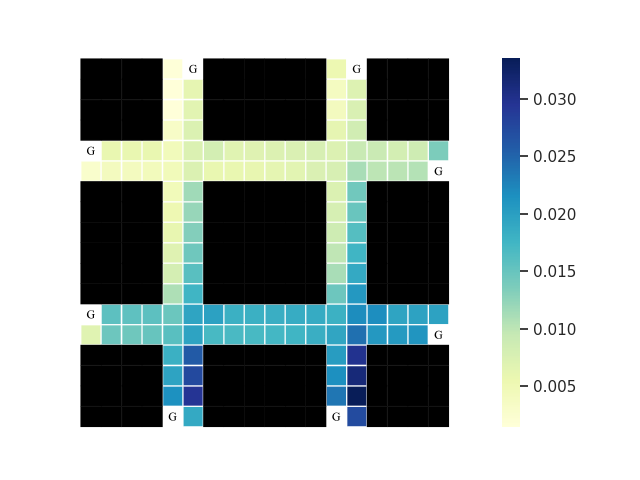}
\caption{Average message vector norm at each position. The eight white cells represent goal positions. It can be seen that the message sent from left lanes are different from that of the right lane. The vector norm might help agent to locate the source of the message.}
\label{fig:norm}
\end{subfigure}

\caption{Analysis of Traffic Junction}
\label{fig:0.1}
\end{figure}

We then increase the $p_{arrive}$ from $0.05$ to $0.1$ and create a harder version of the environment. The implementation follows the procedure described above. Our method achieved $99.3\%$ win rate. IQL with comm and COMA scored $94.3\%$ and $99.1\%$, respectively. We investigate the agent's action selections based on their coordinates. Figure \ref{fig:brake} demonstrates that most brakes occurs when agents are entering junctions or in the grid where vehicles have probabilities to take a turn. Once the agent exit junctions, the number of brakes drops. We also looked at the continuous vectors agents are broadcasting at each location. Following \cite{sukhbaatar2016learning}, we compare the average vector norm of messages sent by agents at each location. As shown in Figure \ref{fig:norm}, the vector norm from opposite lanes are different. The agent might be able to locate the source of the message.

\subsection{Manufacture Line}
\begin{figure}[t]
\begin{subfigure}{0.45\textwidth}
\centering
\includegraphics[width=0.8\linewidth]{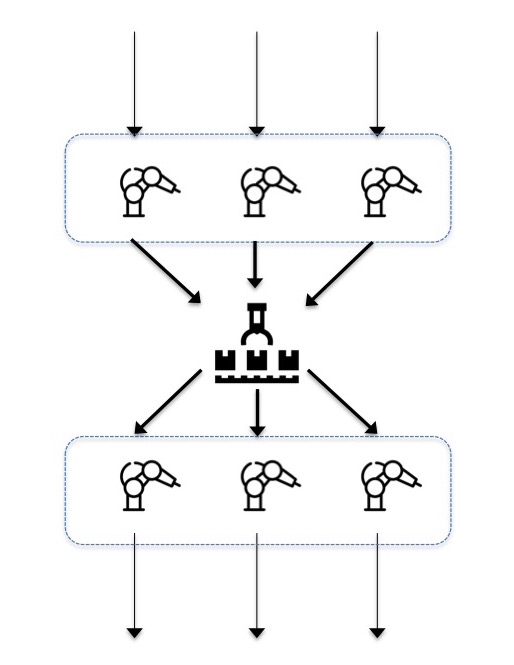} 
\caption{The manufacturing line consists of a 2-step process. The distributor can temporarily store mid-products from step 1 and distribute parts based on the availability of machines in step 2.}
\label{fig:flow}
\end{subfigure}
\hfill
\begin{subfigure}{0.45\textwidth}
\includegraphics[width=1\linewidth]{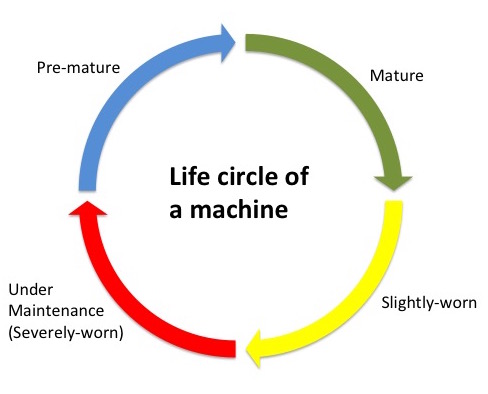}
\caption{The life circle of a machine consists of 4 components. The health state of a machine can start from any point in this life circle e.g. the end of pre-mature, the beginning of severely-worn etc.}
\label{fig:ls}
\end{subfigure}
\caption{Descriptions of manufacturing line environment.}
\label{fig:sm_over}
\end{figure}

The traffic junction environment concerns homogeneous agents whose internal states are deterministic. In addition, the cooperation between agents is not intense in the sense that cars only need to cooperate when they are in the proximity of each other. Inspired by \cite{choo2017health,choo2016adaptive}, we developed this manufacturing line environment which consists of machines of different types and requires constant cooperation between agents. As shown in Figure \ref{fig:flow}, the manufacturing process consists of two steps, where each step contains 3 homogeneous machines. At each timestep, agents take actions, incurring operation costs. The performance of the system heavily depends on the cooperation among machines, e.g. if all machines in step one shut down, machines in step two might as well stop because few parts flows from step 1 to step 2. Otherwise, the system suffers from meaningless operation costs.

The details of the environment are as follows. A product must go through both steps in order to be considered completed. There is a distributor between step 1 and step 2. The distributor can send the partially completed products from an arbitrary machine in step 1 to any machine machine in step 2, and it can temporarily store the partially completed products if process 1 works faster than process 2. Each machine chooses to take actions from \textit{produce}, \textit{stop} and \textit{conducts maintenance}.

Each agent has 4 internal states: pre-mature, mature, slightly-worn, severely-worn. The length of each state is controlled by a gamma distribution of pre-defined mean and scale. This property produces stochastic inner-state transitions for each machine at each life circle. At the beginning of each episode, every machine initializes its internal state randomly. Figure \ref{fig:ls} demonstrates a machine's inner-state transitions. At each timestep, the agent can choose from 3 actions: \textit{produce}, \textit{stop}, and \textit{conduct maintenance}.
If a machine is in the severely-worn state, the only available action is \textit{conduct maintenance}. Otherwise, the machine is free to take any action. If a machine takes the \textit{produce} action, it will incur an operation cost $c_{op}$ and it will be capable of processing $n_{product}$ in this timestep. If a machine takes the \textit{stop} action, it will incur an operation cost $c^{'}_{op}$. If \textit{conduct maintenance} action is selected, the machine will be under maintenance, which incurs maintenance cost $c_{maint}$. The length of maintenance is sampled from a gamma distribution. The only action available during this period is \textit{conduct maintenance}. In addition, cost $c_{broke}$ is incurred if a machine runs to severely-worn state without conducting maintenance in advance. Those rules yield the following reward function at a given timestep $t$:
\begin{equation}
    r(t)=p_{product} * n_t - \sum_{i=1}^{N_t} c_{op} - \sum_{i=1}^{N^{'}_t} c^{'}_{op}- \sum_{i=1}^{\hat{N}_t} c_{maint} - \sum_{i=1}^{\tilde{N}_t} c_{broke}, \nonumber
\end{equation}
where $p_{product}$ is the expected profit of a single product, $n_t$ is the number of products coming out of the 2nd step of the process, $N_t$, $N^{'}_t$, and $\hat{N}_t$ are the total number of machines that take the \textit{produce}, \textit{stop}, and \textit{conduct maintenance} actions respectively, and $\tilde{N_t}$ is the number of machines runs to severely-worn state without conducting maintenance in advance. An episode terminates after 48 steps. Since our reward function is the profit of the system, we evaluate the algorithms by average accumulative profits at the end of test episodes.

\begin{figure}[t]
\centering
\includegraphics[width=.8\textwidth]{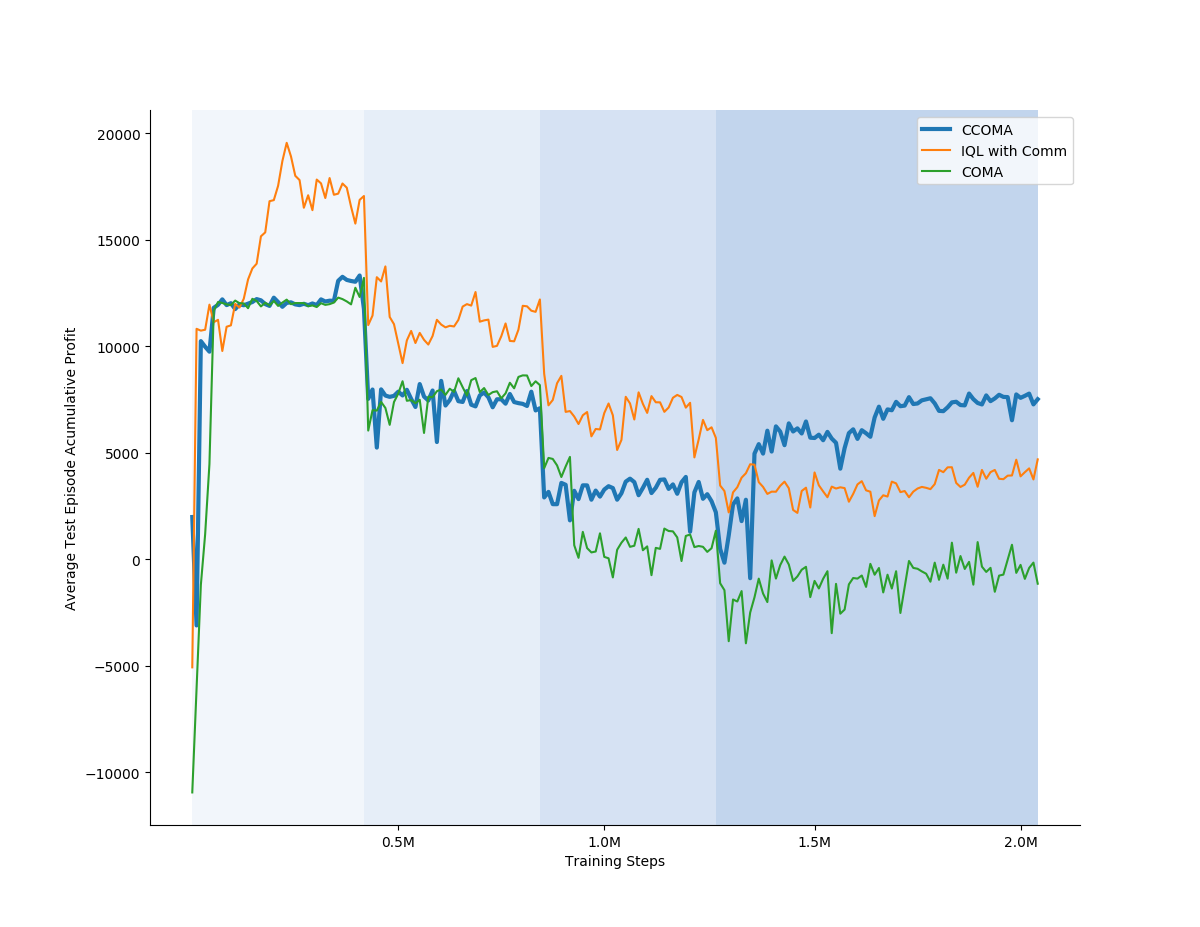}
\caption{Average Test Episode cumulative rewards during Testing episodes. Methods are ran for 5 times and their average performance is presented. The shades corresponds to schedules of curriculum learning.} \label{fig: test_curve}
\end{figure}

In this experiment, each actor is able to observe the internal state of the machine, e.g. health state, machine id, the time has been spent in this state, and the action of the last step. The true state is simply the concatenation of observations from all agents. We utilized curriculum learning \cite{bengio2009curriculum} to train our agents in total $2$ million training steps. Initially, all machines started from the ``brand new''' internal state for the first $0.425$ million training steps. In the following $0.425$ million training steps, $2$ machines were uniformly sampled out of $6$ machines to initialize its internal state randomly at every episode. We continue to sample $2$ more machines every $0.425$ million training steps until all machines initialize their states randomly. The shades in Figure \ref{fig: test_curve} correspond to the schedule of our curriculum learning. As more machines become initialized randomly, the simulation becomes more difficult. 

To understand the effects of the simulation's randomness on the algorithms' learning and performance, we followed the evaluation procedure described in \cite{samvelyan2019starcraft}: the training is paused every $10,000$ steps during which $96$ test episodes are run with the latest model performing greedy action selection. Figure \ref{fig: test_curve} depicts the performance of each method over the training period. It can be seen that the introduction of randomness into the simulation corresponds with a drop in the performance of the algorithms. CCOMA and COMA both show drastic changes in performance when randomness was introduced. However, CCOMA was able to eventually learn good cooperative policies while COMA failed. With the assistance of the centralized critic, CCOMA was able to pick up good policies while IQL with Comm was stuck to sub-optimal policies.  

\section{Conclusion}\label{sec:conclusions}
We introduced CCOMA, a multi-agent RL architecture that allows agents to interact and collaborate. Individually tailored rewards are made possible by a centralized critic that is capable of counterfactual reasoning, with task-specific team reward as the sole feedback in the environment. Furthermore, our investigation in communicated messages illustrated that agents learn meaningful communication strategies under the reward tailoring training paradigm. Evaluations on $2$ diverse tasks clearly show that our architecture is applicable to multi-agent systems with both dynamic sets of agents and a fixed number of agents. Empirically, our methods outperform state-of-art methods from the literature, demonstrating that combining communication with reward shaping is a viable solution to two challenges of MARL. Future work will extend our method to environments with larger numbers of agents. We also aim to implement our framework in real-world applications such as semi-conductor production plants and merging of semi or full self-driving vehicles.

\bibliographystyle{splncs04}
\bibliography{mypaper}

\end{document}